\RequirePackage{fix-cm}
\documentclass[conference]{IEEEtran}
\IEEEoverridecommandlockouts

\usepackage{multirow}
\usepackage{amsmath}
\usepackage{amssymb}
\usepackage[ruled,vlined]{algorithm2e}
\usepackage{graphicx}

\usepackage{xcolor}
\usepackage{colortbl}

\definecolor{publishedbg}{RGB}{237,244,250}

\definecolor{pubhighlight}{RGB}{198,218,239}

\makeatletter
\def\input@path{{IV-2027_EN_2026-11-15/}}
\makeatother
\usepackage[T1]{fontenc}
\usepackage{cite}
\usepackage{microtype}
\usepackage{balance}
\usepackage{amsmath,amssymb,booktabs,array,tabularx}
\usepackage{graphicx}
\usepackage[table]{xcolor}
\usepackage[hidelinks]{hyperref}
\newcommand{\E}{\mathbb{E}}

\DeclareMathOperator*{\argmin}{arg\,min}

\hypersetup{pdftitle={Residual-Guided Lightweight Reinforcement Learning for Risk-Aware Trajectory Selection with a Frozen World Model},pdfauthor={},pdfsubject={Anonymous review manuscript}}
\definecolor{tablegray}{gray}{0.92}
\newcommand{\BaseLtwo}{0.405}
\newcommand{\BaseCR}{0.099}
\newcommand{\FinalLtwo}{0.389}
\newcommand{\FinalCR}{0.098}
\newcommand{\LtwoGain}{3.99}

\newcommand{\LtwoLow}{-0.024}
\newcommand{\LtwoHigh}{-0.009}
\newcommand{\CRLow}{-0.014}
\newcommand{\CRHigh}{0.011}

\newcommand{\ResLtwoGain}{0.8}

\newcommand{\LoRAParamCut}{65.5}

\graphicspath{{IV-2027_EN_2026-11-15/}{./}}
\newenvironment{IEEEkeywords}{\begin{keywords}}{\end{keywords}}

\title{\LARGE \bf Residual-Guided Lightweight Reinforcement Learning for\\Risk-Aware Trajectory Selection with a Frozen World Model}
\author{Anonymous Authors}
\begin{document}
\author{%
Jiaqi Wu$^{1,\dagger}$,
Yongqi Dong$^{1,\dagger,*}$,
Yijie Zhong$^{1}$,
Maple Shao$^{1}$,
Yide Tao$^{2}$,
Linsen Cai$^{1}$,
Weize Cai$^{1}$,\\
Hai L. Vu$^{2}$ and
Alvaro Garc\'ia Hernandez$^{1}$%
\thanks{This work was funded under the Excellence Strategy
of the Federal Government and the L\"ander and with support
from the Exploratory Research Space (ERS) of RWTH.}%
\thanks{$^{1}$Jiaqi Wu, Yongqi Dong, Yijie Zhong, Maple Shao,
Linsen Cai, Weize Cai and Alvaro Garc\'ia Hernandez are with
RWTH Aachen University, Aachen, Germany.}%
\thanks{$^{2}$Yide Tao and Hai L. Vu are with Monash University,
Melbourne, Australia.}%
\thanks{$^{*}$Corresponding author and project leader:
Yongqi Dong.}%
\thanks{$^{\dagger}$Equal contribution.}%
}
\maketitle
\thispagestyle{empty}
\pagestyle{empty}
\begin{abstract}
In automated driving, world models provide future scene predictions for evaluating candidate trajectories, but fixed selection rules may not capture scene-dependent trade-offs. We propose a residual-guided reinforcement learning (RL) framework for lightweight trajectory selection with a frozen world model and a fixed candidate bank. The framework uses proximal policy optimization (PPO) to fine-tune a joint policy over candidate selection and bounded trajectory residuals. A context-conditioned ranking network learns candidate preferences from feedback obtained by evaluating local perturbations of the selected trajectories. At inference, the selector combines learned preference and predicted risk ranks using fixed weights and returns the stored coordinates of the selected candidate trajectory. Fine-tuning updates 16,106 parameters across the actor and critic, while the deployed selector retains only the 11,725-parameter ranking network, with the residual branch and critic used exclusively during training. In open-loop evaluation on the nuScenes validation set, the method reduces average Euclidean trajectory error (L2) by \LtwoGain\% relative to the RiskWorld model baseline without RL-based selection. Under matched budgets for feedback accesses and actor updates, residual exploration reduces L2 by 0.83\% relative to the categorical PPO variant trained without residuals. The method also maintains lower L2 than RiskWorld when risk features are removed or shuffled.

\end{abstract}

\section{Introduction}
In the application of automated driving, world models predict the consequences of candidate actions. 
RiskWorld~\cite{riskworld}, for example, combines occupancy evolution with structured risk representations. Selecting a trajectory from these predictions still requires balancing risk, road geometry, as well as the motion state of the ego vehicle and surrounding vehicles. Handcrafted thresholds and lexicographic priorities specify fixed trade-offs in advance, which can limit adaptation to the scene-dependent candidate preferences.

Existing methods address candidate generation and selection in different ways. SparseDrive~\cite{sparsedrive} uses collision-aware rescoring, while DiffusionDrive~\cite{diffusiondrive} generates multimodal candidates through truncated diffusion. Li \emph{et al.}~\cite{scoring} train scorers over frozen planners using perturbations of logged human trajectories, whereas WorldRFT~\cite{worldrft} applies reinforcement learning (RL) to fine-tune continuous trajectory predictions. A remaining challenge is to train a lightweight selector through learned local exploration around the planner's own candidates, while keeping the world model frozen and returning a preferred candidate at inference.

In this paper, we propose residual-guided RL to train a lightweight selector using feedback from local trajectory variations. A context-conditioned ranking network combines candidate features with pooled candidate-set information to learn preferences and risk coefficients. During training, a joint policy samples a candidate and a bounded residual. Proximal policy optimization (PPO)~\cite{ppo} learns from feedback on the resulting trajectory, evaluated against logged futures using accuracy, progress, smoothness, and safety costs. At inference, the residual head and critic are removed, and a fixed weighted sum of preference and risk ranks selects an original candidate. The main contributions are as follows:
\begin{itemize}
\item A lightweight decision layer learns context-dependent candidate preferences while keeping the world model and candidate generator frozen. Deployment adds only an 11,725-parameter ranking network.
\item A joint candidate-residual policy learns from local trajectory variations during training while preserving candidate geometry at inference. Fine-tuning updates only 16,106 actor and critic parameters in total.
\item Open-loop evaluation on the nuScenes validation set shows a \LtwoGain\% reduction in average L2 relative to the RiskWorld baseline~\cite{riskworld}, together with a lower collision rate. At matched feedback and update budgets, residual exploration further reduces L2 by 0.83\% relative to categorical PPO without residuals. The method also retains lower L2 than RiskWorld when risk features are zeroed or shuffled.
\end{itemize}

\begin{figure*}[t]
\centering\includegraphics[width=0.8599\textwidth]{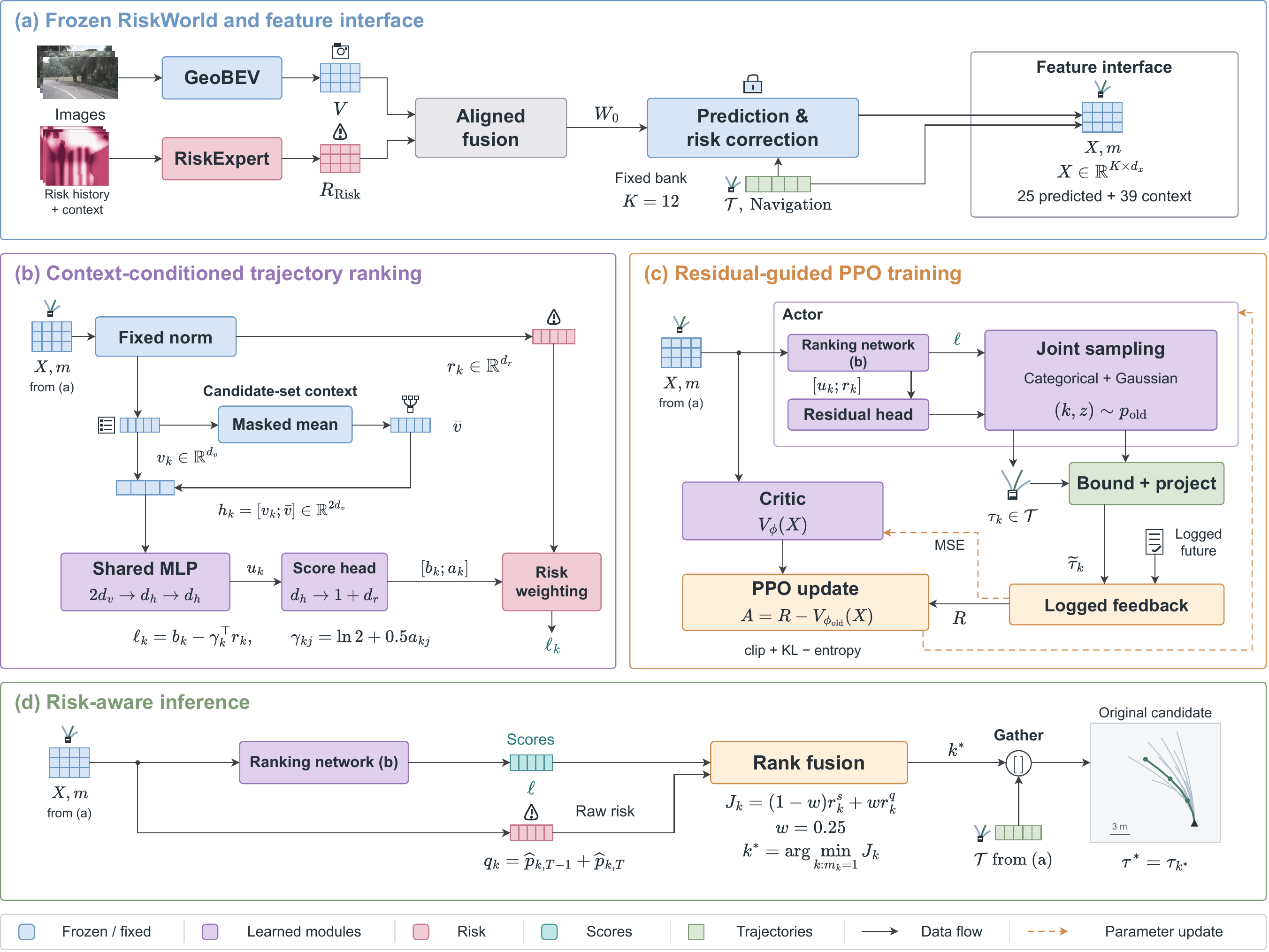}
\caption{Overview of residual-guided reinforcement learning. (a) The frozen RiskWorld~\cite{riskworld} backbone fuses the \textit{RiskExpert} and BEV features in GeoBEV~\cite{geobev}. (b) The ranking network predicts candidate scores. (c) Joint candidate--residual sampling provides feedback for PPO training. (d) Rank fusion selects an original candidate at inference. Training and inference share the ranking network in (b). MSE: mean squared error; KL: Kullback--Leibler divergence.}
\label{fig:architecture}
\end{figure*}


\section{Related Work}
\subsection{World Models and Candidate Planning}
World models inform driving decisions through future scene prediction. DriveDreamer~\cite{drivedreamer} generates driving videos and predicts actions from real-world scenes. RiskWorld~\cite{riskworld} combines shared occupancy forecasting and structured risk representations for selective trajectory intervention. ResWorld~\cite{resworld} models dynamic regions through temporal residuals and refines trajectories using predicted future representations.

For multimodal planning, SparseDrive~\cite{sparsedrive} combines sparse scene representations with collision-aware candidate rescoring. DiffusionDrive~\cite{diffusiondrive} uses multimodal anchors and truncated diffusion for efficient candidate generation. Li \emph{et al.}~\cite{scoring} train scorers over frozen generative planners using lateral and longitudinal perturbations of logged human trajectories.

\subsection{Learnable Costs and Decision Representations}
Trajectory evaluation requires representations of motion preferences and driving costs. Differentiable integrated prediction and planning (DIPP)~\cite{dipp} jointly trains motion prediction and cost weights through a differentiable nonlinear planner. Deep Sets~\cite{deepsets} combines shared element-wise mappings with symmetric aggregation for permutation-invariant representations and permutation-equivariant outputs. Weighted rank aggregation~\cite{rank} integrates ordered preferences from multiple classifier predictions.

\subsection{Policy Optimization and Driving Evaluation}
RL has been used to refine existing controllers and fine-tune trajectory planners. Residual RL~\cite{residualrl} combines a conventional feedback controller with a learned correction to the executed control signal. PPO~\cite{ppo} uses a clipped probability-ratio objective to discourage large policy updates. For automated driving, WorldRFT~\cite{worldrft} fine-tunes continuous trajectory predictions using Gaussianization, collision-aware rewards, and group-based policy optimization. DiffusionDriveV2~\cite{diffusiondrivev2} combines scale-adaptive exploration with intra-anchor and inter-anchor policy updates for multimodal diffusion planning. Planning performance is commonly assessed using open-loop evaluation on the nuScenes~\cite{nuscenes,stp3,vad} benchmark, reporting trajectory displacement and collision metrics. 

\medskip

These aforementioned studies improve future scene prediction, trajectory generation, and learned scoring. 
However, the benefit of policy-guided local exploration for a compact selector with a frozen prediction stack and fixed candidate bank remains less clear. In this paper, we investigate this gap with residual-guided PPO, using feedback from local variations of policy-selected candidates to learn context-dependent candidate preferences.

\begin{table*}[t]
\caption{Evaluation and comparison of Development-stage open-loop planning performance on nuScenes.}
\label{tab:main}\centering
\fontsize{8}{9.6}\selectfont\setlength{\tabcolsep}{2pt}
\begin{tabular*}{\textwidth}{@{\extracolsep{\fill}}lllcrrrrrrrr@{}}
\toprule[0.8pt]
\multirow{2}{*}{Method} & \multirow{2}{*}{Input} & \multirow{2}{*}{Aux. Sup.} & \multirow{2}{*}{\shortstack{Params.\\(M)}} & \multicolumn{4}{c}{L2 (m)$\downarrow$} & \multicolumn{4}{c}{CR (\%)$\downarrow$}\\
\cmidrule(lr){5-8}\cmidrule(l){9-12}
 & & & & 1\,s & 2\,s & 3\,s & Avg. & 1\,s & 2\,s & 3\,s & Avg.\\
\midrule[0.4pt]
\rowcolor{publishedbg} UniAD~\cite{uniad} & C & M+B+Mo+Tr+O & 125.00 & 0.44 & \cellcolor{pubhighlight}0.67 & \cellcolor{pubhighlight}\underline{0.96} & \cellcolor{pubhighlight}0.69 & 0.04 & 0.08 & 0.23 & 0.12\\
\rowcolor{publishedbg}VAD~\cite{vad} & C & M+B+Mo & 58.36 & \cellcolor{pubhighlight}0.41 & 0.70 & \cellcolor{pubhighlight}1.05 & 0.72 & 0.07 & 0.17 & 0.41 & 0.22\\
\rowcolor{publishedbg}SparseDrive-S~\cite{sparsedrive} & C & M+B+Mo+Tr & 85.90 & \cellcolor{pubhighlight}\underline{0.29} & \cellcolor{pubhighlight}\underline{0.58} & \cellcolor{pubhighlight}\underline{0.96} & \cellcolor{pubhighlight}\underline{0.61} & 0.01 & 0.05 & 0.18 & \cellcolor{pubhighlight}\textbf{0.08}\\
\rowcolor{publishedbg}DiffusionDrive~\cite{diffusiondrive} & C & M+B+Mo+Tr & 92.04 & \cellcolor{pubhighlight}\textbf{0.27} & \cellcolor{pubhighlight}\textbf{0.54} & \cellcolor{pubhighlight}\textbf{0.90} & \cellcolor{pubhighlight}\textbf{0.57} & 0.03 & 0.05 & 0.16 & \cellcolor{pubhighlight}\textbf{0.08}\\
\midrule[0.4pt]
RiskWorld~\cite{riskworld} & C+M+GT-Tr & Risk/O+Coll. & 91.099 & 0.182 & 0.374 & 0.660 & 0.405 & 0.068 & 0.083 & 0.147 & 0.099\\
RiskWorld + ER & C+M+GT-Tr & Risk/O+Coll. & 91.110 & 0.179 & 0.369 & 0.653 & 0.400 & 0.068 & 0.083 & 0.156 & 0.103\\
RiskWorld + PPO~\cite{ppo} & C+M+GT-Tr & Risk/O+Coll. & 91.110 & \cellcolor[gray]{0.94}\textbf{0.176} & \cellcolor[gray]{0.94}\underline{0.360} & \cellcolor[gray]{0.94}0.636 & \cellcolor[gray]{0.94}0.391 & 0.068 & 0.083 & 0.150 & 0.100\\
RiskWorld + A3C~\cite{a3c} & C+M+GT-Tr & Risk/O+Coll. & 91.110 & \cellcolor[gray]{0.94}0.178 & 0.363 & 0.641 & 0.394 & 0.088 & 0.106 & 0.158 & 0.117\\
\midrule[0.4pt]
Ours w/o Residual & C+M+GT-Tr & Risk/O+Coll. & 91.110 & \cellcolor[gray]{0.94}\underline{0.177} & \cellcolor[gray]{0.94}0.362 & 0.639 & 0.393 & 0.068 & 0.083 & 0.152 & 0.101\\
Ours with LoRA$^{\dagger}$ & C+M+GT-Tr & Risk/O+Coll. & 91.111 & \cellcolor[gray]{0.94}\textbf{0.176} & \cellcolor[gray]{0.94}\underline{0.360} & \cellcolor[gray]{0.94}\underline{0.634} & \cellcolor[gray]{0.94}\underline{0.390} & 0.068 & 0.083 & 0.143 & \cellcolor[gray]{0.94}\textbf{0.098}\\
\textbf{Residual-guided PPO (ours)}$^{\dagger}$ & C+M+GT-Tr & Risk/O+Coll. & 91.110 & \cellcolor[gray]{0.94}\textbf{0.176} & \cellcolor[gray]{0.94}\textbf{0.359} & \cellcolor[gray]{0.94}\textbf{0.633} & \cellcolor[gray]{0.94}\textbf{0.389} & 0.068 & 0.083 & 0.142 & \cellcolor[gray]{0.94}\textbf{0.098}\\
\bottomrule[0.8pt]
\end{tabular*}
\par\vspace{3pt}\begin{minipage}{\textwidth}\scriptsize
C: cameras; M: maps; GT-Tr: ground-truth actor states. B: 3D object bounding-box; Mo: motion; Tr: tracks; O: occupancy; Coll.: collisions. Aux. Sup. excludes ego-trajectory supervision. Params.: inference parameters (millions).
Light blue backgrounds denote results in published studies with camera-only inputs, differing from those of the local experiments. RiskWorld~\cite{riskworld} is included as the fixed official reference, and all RL-based methods share the frozen prediction stack and candidate set. L2 rankings are computed separately within each group: darker blue and gray shading mark the 3 lowest distinct values in each column for the published and local results, respectively. Bold and underlining indicate the lowest and second-lowest values. For CR, only the lowest average value within each group is highlighted. Ties after rounding share formatting. $\dagger$: training-time residual exploration.




\end{minipage}

\end{table*}

\begin{table*}[t]
\caption{Planning performance under inference-time perturbations of RiskExpert features.}
\label{tab:perturbation}\centering\footnotesize
\begin{tabular*}{\textwidth}{@{\extracolsep{\fill}}lrrrr@{}}
\toprule[0.8pt]
\multirow{2}{*}{Risk features} & \multicolumn{2}{c}{RiskWorld~\cite{riskworld}} & \multicolumn{2}{c}{Ours}\\
\cmidrule(lr){2-3}\cmidrule(l){4-5}
 & L2 (m)$\downarrow$ & CR (\%)$\downarrow$ & L2 (m)$\downarrow$ & CR (\%)$\downarrow$\\
\midrule[0.4pt]
Original (Full) & 0.4054 & 0.0993 & \cellcolor[gray]{0.94}\textbf{0.3893} & \cellcolor[gray]{0.94}\textbf{0.0979}\\
Zeroed & 0.6399 & 0.2816 & \cellcolor[gray]{0.94}\textbf{0.5698} & \cellcolor[gray]{0.94}\textbf{0.2491}\\
Shuffled & 0.5004 & 0.2675 & \cellcolor[gray]{0.94}\textbf{0.4636} & \cellcolor[gray]{0.94}\textbf{0.2344}\\
\bottomrule[0.8pt]
\end{tabular*}
\par\vspace{3pt}\begin{minipage}{\textwidth}\scriptsize
Metrics are averaged over 1, 2, and 3\,s. The Original RiskWorld row uses the fixed official reference. Shaded bold entries mark the lower value in each row.
\end{minipage}

\end{table*}

\begin{figure*}[t]
\centering
\includegraphics[width=0.68636\textwidth]{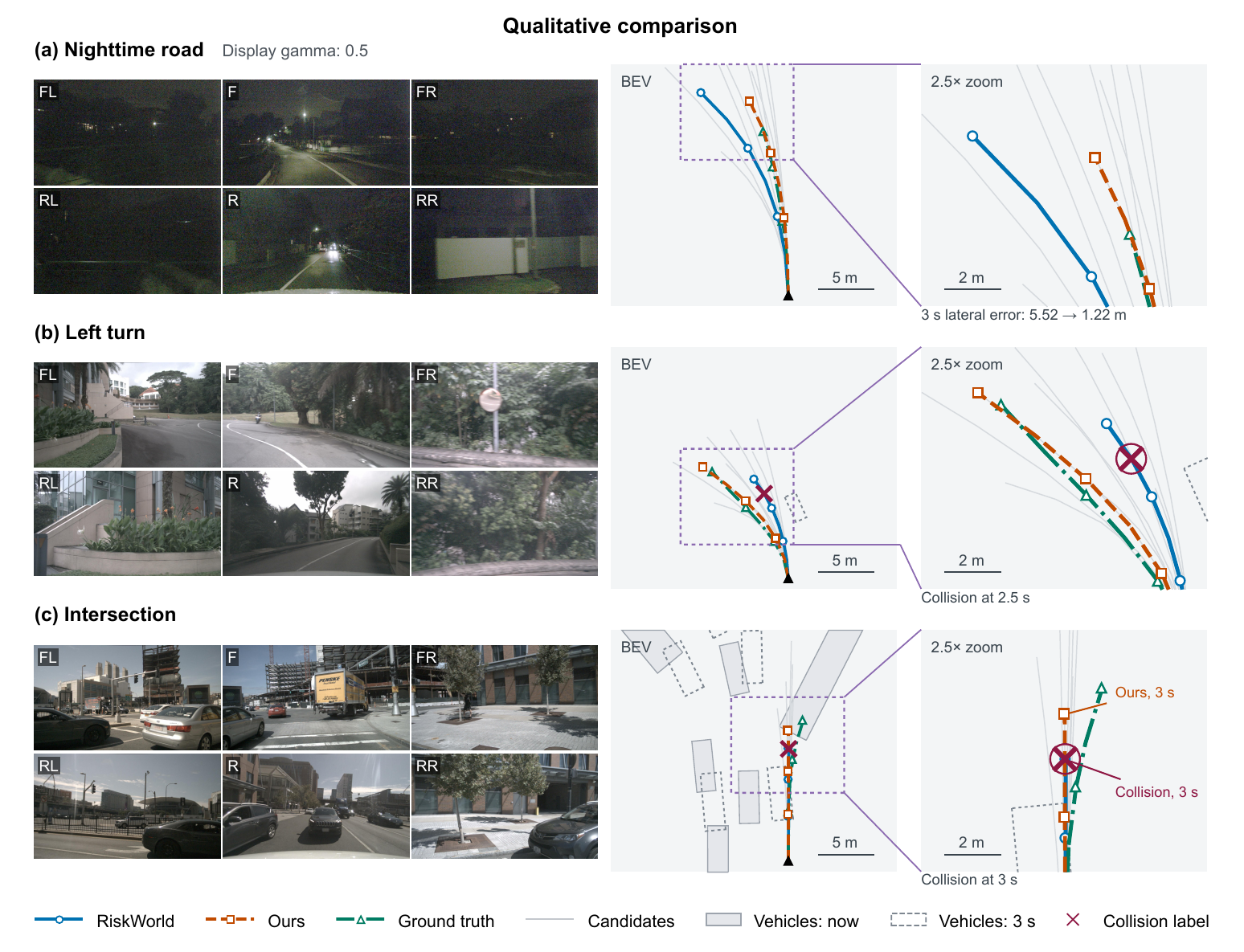}
\caption{Qualitative comparison on nuScenes: (a) nighttime road, (b) left turn, and (c) intersection. Each row shows six camera views, BEV trajectories, and a close-up. Crosses mark collisions; trajectory markers indicate 1, 2, and 3\,s. FL/F/FR and RL/R/RR denote the front-left/front/front-right and rear-left/rear/rear-right cameras. Nighttime views are gamma-corrected with an exponent of 0.5 for visualization.}
\label{fig:qualitative}
\end{figure*}

\section{Method}
\label{sec:method}

\textbf{Fig.~\ref{fig:architecture}} shows the residual-guided RL framework. Frozen RiskWorld~\cite{riskworld} fuses the \textit{RiskExpert} representation and the bird's-eye-view (BEV) features from GeoBEV~\cite{geobev} into candidate-aligned inputs. A context-conditioned trajectory ranking network learns through PPO with joint candidate-residual feedback. At inference, rank fusion combines learned-score and predicted-risk ranks to select an original candidate. RiskWorld remains frozen throughout; residual exploration is confined to training.

\subsection{Decision Learning over Frozen Predictions}
For each scene context $i$, frozen RiskWorld provides $K=12$ trajectories $\{\tau_{ik}\}_{k=1}^{K}$, validity masks $m_i\in\{0,1\}^{K}$, and aligned features $X_i\in\mathbb R^{K\times d_x}$. Each trajectory possesses $T=6$ states at 0.5\,s intervals. The $d_x=64$ features comprise 25 predictions and 39 state/navigation descriptors covering collisions, road risks, clearance, motion, and navigation. The prediction stack and candidate generator remain fixed.

We formulate adaptation as a one-step contextual decision. Given $X_i$ and $m_i$, the actor samples a candidate and a conditional residual. An evaluator scores the resulting trajectory against logged futures, with surrounding actors following their recorded trajectories. The actor's ranking network and residual head share features, while a separate critic estimates the context value. Only the actor and critic receive gradients.

\textbf{Fig.~\ref{fig:architecture}} omits the context index $i$. In panel (a), the RiskExpert representation $R_{\mathrm{Risk}}$ and GeoBEV features $V$ are fused as $W_0=R_{\mathrm{Risk}}+G\odot\Delta$, where $\Delta$ is a learned feature correction computed from aligned visual features $V$, risk features $R_{\mathrm{Risk}}$, and their coverage mask, and $G$ is a coverage-dependent gate. 

\subsection{Context-Conditioned Trajectory Ranking}
Ranking scores define the categorical training policy and enter rank fusion at inference. Features are standardized with fixed training statistics and split into $r_{ik}\in\mathbb R^{d_r}$ and $v_{ik}\in\mathbb R^{d_v}$. Here, $d_r=2T$ covers collision/off-road predictions at each step, while $d_v=d_x-d_r$ covers the remaining predictions and state/navigation features. To condition scores on the available alternatives, we concatenate $v_{ik}$ with its mean over valid candidates:
\begin{equation}
 \bar v_i=\frac{\sum_k m_{ik}v_{ik}}{\sum_k m_{ik}},\qquad
 h_{ik}=[v_{ik};\bar v_i]\in\mathbb R^{2d_v}.
 \label{eq:context}
\end{equation}
Every context has a valid candidate. A shared multilayer perceptron (MLP) with two width-$d_h=64$ hidden layers and rectified linear unit (ReLU) activations maps $h_{ik}$ to $u_{ik}\in\mathbb R^{d_h}$. The score head in \textbf{Fig.~\ref{fig:architecture}} (b) produces $1+d_r$ outputs: a base preference $b_{ik}\in\mathbb R$ and risk-weight offsets $a_{ik}\in\mathbb R^{d_r}$. The score is
\begin{equation}
 \ell_{ik}=b_{ik}-\gamma_{ik}^{\top}r_{ik},\quad
 \gamma_{ikj}=\ln 2+0.5a_{ikj},
 \label{eq:score}
\end{equation}
where $j=1,\ldots,d_r$ indexes the risk features. The learned coefficients $\gamma_{ikj}$ act on standardized risk features and may be negative. Mean pooling is permutation-invariant. The shared mapping is permutation-equivariant, so reordering candidates reorders their scores. A masked softmax defines $\pi_\theta(k\mid X_i)$, assigning zero probability to invalid candidates.


\subsection{Joint Learning with Local Residual Feedback}
Local residual exploration provides training feedback from trajectories near the selected candidate. An affine residual head maps $[u_{ik};r_{ik}]\in\mathbb R^{d_h+d_r}$ to the mean $\mu_{ik}\in\mathbb R^2$ of a diagonal Gaussian over the latent $z\in\mathbb R^2$. The learned log-scale vector $\log\sigma\in\mathbb R^2$ is shared across candidates and scenes and clamped to $[-3,0.5]$. The joint policy is
\begin{equation}
p_\theta(k,z\mid X_i)
=\pi_\theta(k\mid X_i)
\mathcal N\!\left(z;\mu_{ik},\operatorname{diag}(\sigma^2)\right).
\label{eq:joint}
\end{equation}
The residual head's dependence on $u_{ik}$ allows residual feedback to update the shared ranking features through the joint objective.

The sampled latent defines a bounded displacement $d=(1.0,0.5)\odot\tanh z$ (m), applied along the trajectory using the smoothstep weight $3s_t^2-2s_t^3$, where $s_t=t/T$. If the perturbed trajectory violates acceleration, curvature, or forward-motion constraints, a geometric projection progressively shrinks the displacement. Original candidate is used if no shrinkage step satisfies these constraints. The evaluator combines trajectory accuracy, progress, and smoothness with safety and road costs:
\begin{equation}
U=-e+\lambda_P P-\lambda_C C,\qquad
R=U-\omega^\top c.
\label{eq:reward}
\end{equation}
Here, $e$ denotes trajectory error, $P$ is a progress proxy, and $C$ is a motion-smoothness cost. The four components of $c$ are expressed as fractions and quantify collision exposure, any-collision events, road violations, and mean collision exposure under small translations. We fix $\lambda_P=\lambda_C=10^{-3}$ and $\omega=(1500,20,3,500)^\top$.

Each context supplies one valid candidate and conditional residual sampled from the old joint policy. The separate critic $V_\phi(X_i)$ predicts a scalar value from the $d_x$-dimensional mean of normalized valid-candidate features using an MLP with one width-$d_h$ hidden layer. Its pre-update prediction defines the advantage:
\begin{equation}
 A_i=R_i-V_{\phi_{\mathrm{old}}}(X_i),\qquad
 \mathcal L_V=\tfrac12\E_i\big[(V_\phi(X_i)-R_i)^2\big].
 \label{eq:advantage}
\end{equation}
Advantages are standardized within each batch as $\widehat A_i=(A_i-\bar A)/\max(\operatorname{std}(A),10^{-6})$, using the population standard deviation, and remain fixed during the update. The critic is trained with detached inputs and a separate optimizer, so value regression updates its own parameters.

PPO~\cite{ppo} updates the actor using the joint probability ratio $\rho_i=p_\theta(k_i,z_i\mid X_i)/p_{\mathrm{old}}(k_i,z_i\mid X_i)$:
\begin{equation}
 \begin{aligned}
 \mathcal L_\pi={}&-\E_i\big[\min\{\rho_i\widehat A_i,
 \operatorname{clip}(\rho_i,1-\epsilon,1+\epsilon)\widehat A_i\}\big]\\
 &+\beta\E_i\mathcal K_i-\eta\E_i\mathcal H(\pi_\theta).
 \end{aligned}
 \label{eq:ppo}
\end{equation}
The regularizer $\mathcal K_i=D_{\mathrm{KL}}(p_\theta\|p_{\mathrm{ref}})$ penalizes deviations from fixed reference policy, and $\mathcal H(\pi_\theta)$ denotes categorical entropy. The probability ratio is evaluated in latent space for the sampled candidate and latent $z$ under the current and old joint policies. The Jacobian of the shared $\tanh$ transformation cancels in this ratio, while geometric projection affects only feedback evaluation. At inference, no residual is sampled or applied, and neither the critic nor the feedback evaluator is used.

\subsection{Risk-Aware Selection at Inference}
\label{sec:inference}
The selector combines learned preference with raw collision predictions from the frozen world model. The risk signal is $q_{ik}=\widehat p_{ik,T-1}+\widehat p_{ik,T}$, where $\widehat p_{ik,t}$ is the predicted collision probability for candidate $k$ at future step $t$. Among valid candidates, scores and risks are converted to ranks $r^s_{ik}=\operatorname{rank}_{\downarrow}(\ell_{ik})$ and $r^q_{ik}=\operatorname{rank}_{\uparrow}(q_{ik})$. Selection minimizes their weighted sum:
\begin{equation}
 \begin{aligned}
 J_{ik}&=(1-w)r^s_{ik}+wr^q_{ik},\\
 k_i^*&=\argmin_{k:m_{ik}=1}J_{ik},\qquad \tau_i^*=\tau_{i,k_i^*}.
 \end{aligned}
 \label{eq:rank}
\end{equation}


Scores and risks are mapped to a common ordinal scale using zero-based ranks obtained through two argsort operations. Ties in $J$ are resolved by selecting the lowest candidate index. We select $w$ through out-of-fold (OOF) evaluation on the training scenes, subject to an accuracy constraint. The selected value, $w=0.25$, is fixed for all evaluation scenes (\textbf{Fig.~\ref{fig:architecture}}(d)). The output trajectory retains the selected candidate's coordinates.

The actor contains 11,881 parameters, including 156 in the residual head, and the critic contains 4,225, giving a total of 16,106 trainable parameters during fine-tuning. At inference, the residual head and critic are removed, leaving the 11,725-parameter ranking network.

\section{Experiments and Results}
In this section, we evaluate the proposed selector against the fixed selection rule in RiskWorld~\cite{riskworld}, assess the contribution of residual exploration under matched feedback and update budgets, and examine sensitivity to risk-feature perturbations.

\subsection{Dataset and Metrics}
Training and development use 698 scenes from nuScenes~\cite{nuscenes}, split by scene into 18,989 fitting contexts and 4,779 internal-validation contexts. Open-loop evaluation uses 5,119 contexts with complete future annotations from the 150 nuScenes validation scenes, all excluded from training. Inputs comprise annotated actor states and maps; feedback is derived from logged future trajectories. We use the fixed official RiskWorld~\cite{riskworld} results as the reference. All learning methods in \textbf{Table~\ref{tab:main}} share the same frozen predictions and $K=12$ candidates. Published results with camera-only inputs~\cite{uniad, vad, sparsedrive, diffusiondrive} are included only for complementary reference because their input settings differ from ours.

Following VAD~\cite{vad} and ST-P3~\cite{stp3}, we report Euclidean trajectory error (L2) and collision rate (CR, \%), with collisions assessed against annotated agent boxes. At each horizon (1, 2, and 3\,s), both metrics are averaged over all preceding time steps, including the horizon itself; \textit{Avg.} is the mean of these three values. We average learning-based results over three training runs with matched random seeds across methods. We estimate 95\% confidence intervals (CIs) with 2,000 paired scene-bootstrap resamples, preserving within-scene dependence. These intervals characterise variability across evaluation scenes conditional on the trained models. Relative changes are calculated from unrounded metric values.


\subsection{Implementation Details}
All learning-based methods are initialized from the same ranking-network checkpoint pretrained with RL feedback. The reference policy $p_{\mathrm{ref}}$ in~\eqref{eq:ppo} is held fixed at its initial parameters. All methods share feature normalization, candidate trajectories, the reward function, and the inference weight $w=0.25$. Each method is fine-tuned independently and evaluated after its final update. We use AdamW~\cite{adamw} as the optimizer, with the learning rate and weight decay both set to $10^{-4}$. The PPO clipping parameter $\epsilon$, KL weight $\beta$, and entropy weight $\eta$ are set to 0.2, 0.1, and 0.01, respectively.

\textbf{Learning objectives.} During fine-tuning, ER, PPO, and A3C use feedback only from original candidates. Expected return (ER) enumerates valid candidates and maximizes their policy-weighted cached return without sampling actions. PPO~\cite{ppo} and four-worker asynchronous advantage actor--critic (A3C)~\cite{a3c} sample candidates and use critics. Our method and its low-rank adaptation (LoRA)~\cite{lora} variant also explore bounded residuals, as indicated by $\dagger$ in \textbf{Table~\ref{tab:main}}. LoRA freezes the ranking-network weights and trains rank-4 adapters on both hidden mappings, while retaining joint PPO, the critic, and residual exploration. Ours w/o Residual uses categorical PPO without residual exploration during fine-tuning.

\textbf{Training budgets.} Each cached-return read or residual-trajectory evaluation counts as one feedback access. Our method, its LoRA variant, and Ours w/o Residual each use 9,830,400 feedback accesses and 800 actor updates per run. ER uses 2,457,600 accesses per run; PPO and A3C each use 204,800 accesses and 800 actor updates. The residual and LoRA ablations use matched feedback and update budgets. Comparisons with ER, PPO, and A3C reflect differences in both the learning objective and feedback budget.

\subsection{Main Results}
Compared with the fixed official RiskWorld~\cite{riskworld} baseline, our method reduces average L2 from \BaseLtwo\,m to \FinalLtwo\,m, a relative reduction of \LtwoGain\% (\textbf{Table~\ref{tab:main}}). The largest horizon-level improvement occurs at 3\,s. The paired 95\% CI for the L2 difference (ours minus RiskWorld) is $[\LtwoLow,\LtwoHigh]$\,m. The average CR point estimate decreases from \BaseCR\% to \FinalCR\%, with a corresponding 95\% CI of $[\CRLow,\CRHigh]$ percentage points for the difference.


PPO and A3C also reduce average L2 relative to the RiskWorld~\cite{riskworld} baseline, but their CR point estimates of 0.100\% and 0.117\%, respectively, exceed the reference value. Our method achieves lower average L2 and CR point estimates than both baselines under the same frozen-candidate interface, using a slightly larger feedback budget.

\textbf{Fig.~\ref{fig:qualitative}} illustrates how candidate selection changes lateral position, turning curvature, and timing. On the nighttime road (a), absolute lateral-coordinate error at 3\,s decreases from 5.52 to 1.22\,m. In the left-turn scenario (b), our method selects a candidate with adjusted turning curvature; at the intersection (c), it selects one with adjusted forward timing. Evaluated against recorded actor trajectories, these candidates avoid the collisions as in the RiskWorld baseline at 2.5\,s and 3\,s, respectively.

\subsection{Ablation Studies}
\label{sec:ablation}
\textbf{Residual exploration.} Relative to the RiskWorld~\cite{riskworld} baseline, Ours w/o Residual achieves lower average L2 but a higher CR point estimate (0.101\% versus 0.099\%). With matched feedback and update budgets, adding residual exploration further reduces average L2 by \ResLtwoGain\% and lowers the CR point estimate to 0.098\%. Both variants share the same initialization and return original candidate trajectories at inference. During fine-tuning, feedback from residual trajectories also contributes to updates of the shared ranking features.

\textbf{Low-rank adaptation (LoRA).} LoRA uses \LoRAParamCut\% fewer trainable actor and critic parameters than the full method. Its average L2 is 0.001\,m higher, and its average CR matches that of the full method (\textbf{Table~\ref{tab:main}}). Both point estimates remain below the corresponding RiskWorld~\cite{riskworld} reference values.



\subsection{Sensitivity to Risk-Feature Perturbations}
Following RiskWorld~\cite{riskworld}, we hold all model weights fixed during the risk-feature perturbation tests. \textit{Original} (Full) uses unperturbed inputs. \textit{Zeroed} tests reliance on risk information by setting the RiskExpert representation ($R_{\mathrm{Risk}}$ in \textbf{Fig.~\ref{fig:architecture}}(a)) to zero before fusion. \textit{Shuffled} tests scene correspondence by exchanging representations across contexts while preserving each receiving context's validity mask. Both perturbations increase L2 and CR for both methods (\textbf{Table~\ref{tab:perturbation}}). Compared with RiskWorld ~\cite{riskworld}, our method retains lower point estimates for both metrics, with L2 of 0.5698 versus 0.6399\,m under \textit{Zeroed} and 0.4636 versus 0.5004\,m under \textit{Shuffled}.

\section{Conclusion}

This paper presents a residual-guided RL method for lightweight trajectory selection with a frozen world model and candidate generator. Joint candidate-residual sampling provides local trajectory feedback for training a context-conditioned ranker. At inference, the compact ranking network combines learned preference and predicted risk ranks to select an original candidate and return its stored coordinates. On nuScenes, the method reduces average open-loop Euclidean trajectory error (L2) by \LtwoGain\% relative to the fixed official RiskWorld~\cite{riskworld} reference. Matched-budget ablations show a further improvement from residual exploration over categorical PPO without residuals. The advantage regarding L2 error over the RiskWorld~\cite{riskworld} baseline also persists when risk features are zeroed or shuffled.

We evaluate the method in an open-loop setting with a single backbone and dataset, using annotation-derived actor states and maps. Future work should prioritise reactive closed-loop evaluation to assess executed driving performance and interactions with other road users. Evaluation across additional backbones and datasets with perception-derived inputs would assess transfer and robustness to perception errors and distribution shifts. Further work includes adaptive candidate generation for better coverage in challenging scenes and broader evaluation across training runs and rare-event scenarios to characterise collision outcomes.



\bibliographystyle{IV-2027_EN_2026-11-15/IEEEtran}
\bibliography{IV-2027_EN_2026-11-15/references}
\end{document}